\documentclass[11pt]{article}

\usepackage[final]{acl}

\usepackage{times}
\usepackage{latexsym}

\usepackage[T1]{fontenc}

\usepackage[utf8]{inputenc}

\usepackage{microtype}

\usepackage{inconsolata}

\usepackage{graphicx}

\usepackage{amsmath}
\usepackage{amssymb}
\usepackage{enumitem}
\usepackage{hyperref}
\hypersetup{
 colorlinks=true,
 citecolor=cyan,
 linkcolor=BurntOrange,
 urlcolor=cyan}
\usepackage[capitalize,noabbrev]{cleveref}
\usepackage{algorithm}
\usepackage{algpseudocode}
\usepackage{tikz}
\newcommand*\circled[1]{\tikz[baseline=(char.base)]{
            \node[shape=circle,draw,inner sep=0.5pt] (char) {#1};}}
\usepackage[most]{tcolorbox}
\newtcbox{\roundbox}{
  on line,
  colback=gray!20!white,
  colframe=gray!20!white,
  coltext=black!50,
  boxrule=0pt,
  arc=3pt,
  boxsep=1pt,
  left=2pt,
  right=2pt,
  top=1pt,
  bottom=1pt
}
\usepackage{booktabs}
\usepackage{adjustbox}
\usepackage{dsfont}

\newcommand{\graycomment}[1]{{\color[HTML]{C0C0C0}{$\triangleright$ #1}}%
}
\usepackage{listings}

\lstdefinelanguage{json}{
  basicstyle=\ttfamily\footnotesize,
  breaklines=true,
  breakatwhitespace=false,
  columns=fullflexible,
  keepspaces=true,
  showstringspaces=false,
}

\title{SEGRA: A Structured Experience Guided Reasoning Agent \\for Property Graph Question Answering}

\author{
\textbf{Saiyue Lyu\textsuperscript{1}}{\hypersetup{linkcolor=black}\thanks{Corresponding author. Work done during an internship at Amazon.}},
\textbf{Mariam Dundua\textsuperscript{2}},
\textbf{Vishaal Kapoor\textsuperscript{2}},
\textbf{Sarthak Ahuja\textsuperscript{2}},
\\
\textbf{Neda Kordjazi\textsuperscript{2}},
\textbf{Evren Yortucboylu\textsuperscript{2}},
\textbf{Harsh Amin\textsuperscript{2}},
\textbf{Rebecca Steinert\textsuperscript{2}}
\\
 \textsuperscript{1}University of British Columbia,
 \textsuperscript{2}Amazon,
\\
\texttt{saiyue.lyu@ubc.ca},\,\,\texttt{\{madundua,vishaalk\}@amazon.com}
\\
\texttt{\{sarahuja,nedakord,yortuc,haminate,rsteinrt\}@amazon.com}
}

\begin{document}

\maketitle


\begin{abstract}
Enterprise IT support knowledge graphs capture rich relationships among cases, users, devices, symptoms, taxonomic categories, root causes, and historical resolutions. Yet querying them in Gremlin requires knowledge of graph schemas, traversal semantics, edge directionality, and property-graph-specific constraints, making them difficult for non-expert operators to use. We introduce SEGRA, an experience-guided agent for enterprise text-to-Gremlin question answering. SEGRA integrates intent routing, schema- and taxonomy-grounded query generation, multi-shot decomposition, execution-aware verification, and a curriculum-bootstrapped skill library that reuses verified query patterns.
On an enterprise IT support benchmark, SEGRA achieves a $7.0\times$ higher mean judge score than backbone-only chain-of-thought prompting. Its skill library further reduces LLM calls by $20\%$ and dollar cost by $18\%$ relative to SEGRA without skills, while preserving answer quality. These results show that schema-grounded agent design and reusable execution experience improve both accuracy and efficiency for enterprise graph QA.
\end{abstract}

\section{Introduction}


Graph databases are widely used to model connected data in applications such as fraud detection, knowledge management, recommendation, and enterprise analytics \citep{guo2020survey,huang2022dgraph,besta2023demystifying}. Unlike relational databases, they represent information as vertices and edges, allowing connected entities to be traversed directly rather than reconstructed through joins \citep{rodriguez2015gremlin,francis2018cypher}. This structure is well suited to enterprise IT support, where cases are linked to users, devices, symptoms, root causes, resolutions, timestamps, and related historical incidents. However, querying such graphs requires fluency in graph query languages, schema labels, edge directionality, and label-versus-property semantics \citep{angles2017foundations}, creating a barrier for non-expert operators, who must rely on a few specialists to query the graph at production volume.

Large language models offer a natural way to bridge this gap by translating user questions into executable database queries. Text-to-SQL has been extensively studied for relational databases, with mature benchmarks and methods for mapping natural language to SQL \citep{pourreza2023din, chen2024beaver, wang2025mac}. In contrast, text-to-graph-query generation remains less developed. Graph databases introduce challenges beyond SQL, including traversal semantics, schema-specific vertex and edge labels, and label-versus-property distinctions. While Cypher generation has received recent attention \citep{ozsoy2025text2cypher, lyu2026text2gql}, Gremlin generation remains relatively understudied in the academic literature.


Enterprise IT support graphs introduce more challenges beyond standard text-to-database settings. Diagnostic questions often require intermediate operations such as comparisons, set differences, or top-N aggregation, making single-step Gremlin generation brittle. The model must also align user phrasing with a domain taxonomy of symptoms, causes, categories, and resolutions, rather than guessing schema labels from surface text. In addition, derived similarity edges encode domain-specific relationships whose traversal patterns are easy to misuse without explicit guidance. These challenges make enterprise IT support a demanding testbed for natural-language-to-Gremlin systems.

\begin{figure*}
    \centering
\includegraphics[width=0.8\linewidth]{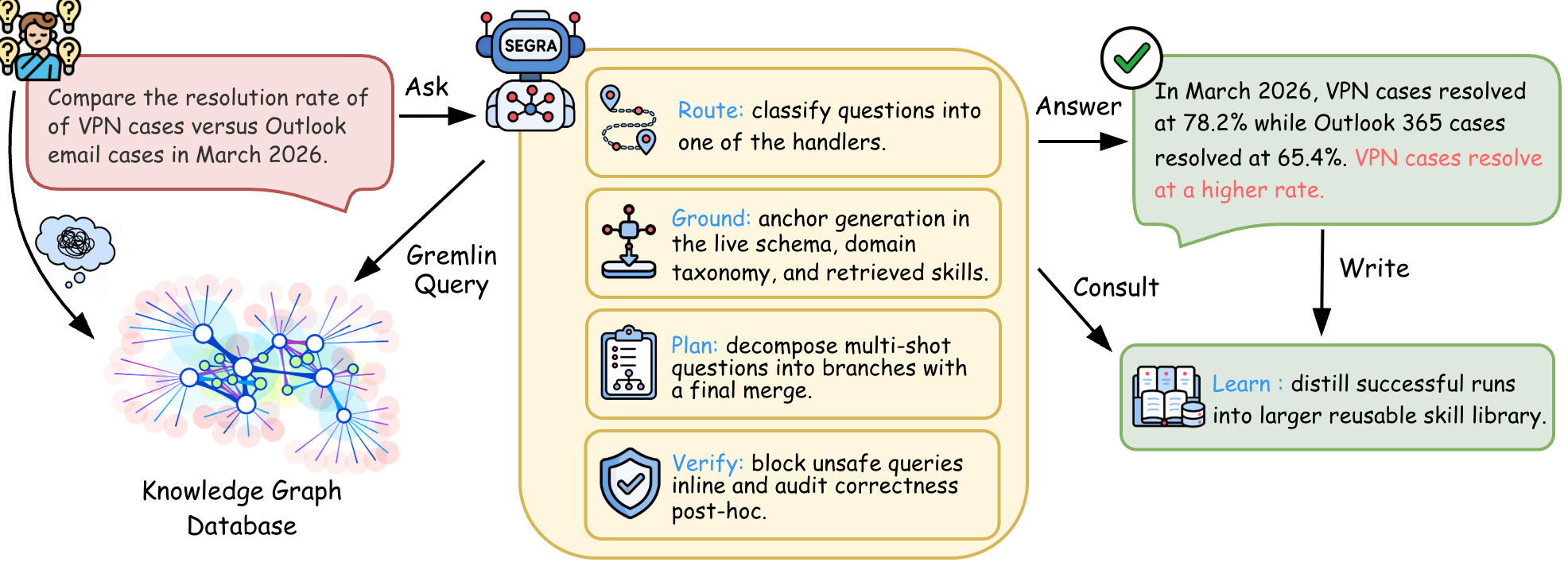}
    \caption{Overview of SEGRA framework. SEGRA takes natural-language questions as input and produces Gremlin queries to further generate natural-language answers.}
    \label{fig:overview}
\end{figure*}

We propose \textbf{SEGRA}, a {\bf S}tructured {\bf E}xperience {\bf G}uided {\bf R}easoning {\bf A}gent for question answering over enterprise property graphs, as illustrated in \cref{fig:overview}. SEGRA grounds generation in the graph schema and taxonomy, supports similarity queries through dedicated traversal patterns, and decomposes complex questions into verified sub-queries. Because public text-to-Gremlin benchmarks remain limited, SEGRA is designed to accumulate experience from successful executions: verifier-passing executions are distilled into reusable skills that capture effective query patterns and reasoning strategies. These skills guide future queries and further reduce LLM calls and inference cost.

SEGRA is a system-level framework that unifies intent routing, schema-grounded generation, similarity handling, multi-shot planning, execution-aware verification, and cross-query skill accumulation. To the best of our knowledge, it is the first text-to-Gremlin agent to pair reusable execution experience with property-graph-specific correctness checks. Although evaluated on enterprise IT support, the design applies to property-graph interfaces requiring domain grounding, derived relations, and multi-step query composition.

\section{Related Work}

\noindent\textbf{Natural-language querying of structured data.}
Text-to-SQL has been extensively studied for relational databases \citep{yu2018spider,li2023can}, while text-to-graph-query generation has received recent attention primarily for Cypher \citep{ozsoy2025text2cypher,lyu2026text2gql}, leaving Gremlin relatively underexplored. KGQA systems combine retrieval and reasoning over entity-relation graphs \citep{sun2024think,luo2024chatkbqa} but target answer retrieval rather than executable queries. SEGRA addresses the less-studied setting of generating executable Gremlin over enterprise property graphs, whose schemas combine domain taxonomies, temporal constraints, and derived similarity edges, and whose queries require  traversal operators, edge directionality, aggregation, and execution constraints.

\noindent\textbf{Experience-guided LLM agents.} Recent LLM agents improve without weight updates by accumulating reusable experience: Voyager \citep{wang2023voyager} stores successful action programs in a skill library, AFlow \citep{zhang2025aflow} searches over workflows of LLM operators, and related systems explore persistent memory and reflection \citep{shinn2023reflexion,zhang2025darwin}. The absence of public Text-to-Gremlin training data makes this direction a natural fit for enterprise property-graph QA: SEGRA reuses execution experience by representing complex questions as workflow DAGs, verifying candidates with deterministic constraints, and distilling verifier-passing runs into reusable skills indexed by question template.

\noindent\textbf{IT support question answering.}
LLM-based IT support work spans diagnostic dialogue \citep{kapoor2026dqa,ahuja2026vigil}, cloud-incident root-cause analysis \citep{ahmed2023recommending,roy2024exploring, chen2025aiopslab}, and natural-language-to-Gremlin generation \citep{yun2025eicopilot}. The closest, EICopilot \citep{yun2025eicopilot}, generates a single Gremlin script per question for entity-centric summarization using a static example bank. SEGRA instead targets analytical questions over a historical support graph -- aggregations, comparisons, top-$N$, set differences, and similarity traversal -- motivating two design choices: multi-shot decomposition with a final merge, since single-shot generation breaks on compositional questions; and an experience-guided skill library that grows online from verifier-passing runs.

\section{Method}
\label{sec:method}

We denote the enterprise property graph by $G$ and its live schema by $\Sigma$ (vertex labels, edge labels, per-label property keys), discovered once at startup. Let $\mathcal{T}$ denote the domain taxonomy of case categories.

\begin{figure*}[t!]
    \centering
    \includegraphics[width=0.9\linewidth]{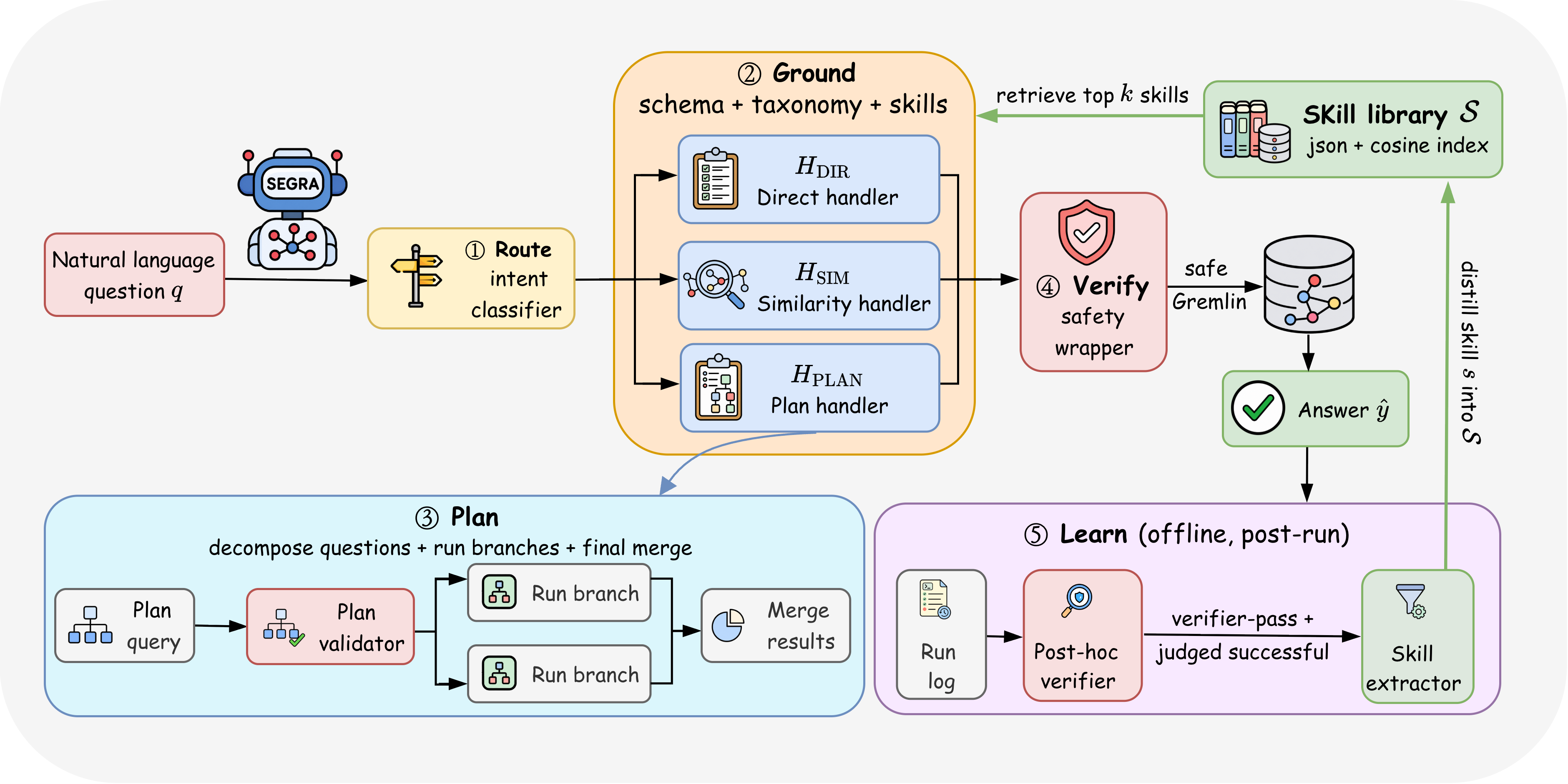}
    \caption{Pipeline of SEGRA framework: route to a handler, ground in the live schema and taxonomy with retrieved skills, optionally plan and execute branches, verify, and distill verifier-passing runs into the skill library that guides future questions.}
    \label{fig:pipeline}
\end{figure*}

As shown in \cref{fig:pipeline}, SEGRA processes a natural-language question $q$ in five stages.
\circled{1} \textbf{Route.} An intent classifier assigns $q$ a route label $\rho \in \mathcal{R} = \{\textsc{dir}, \textsc{sim}, \textsc{plan}\}$, each corresponding to a handler.
\circled{2} \textbf{Ground.} SEGRA grounds generation in $\Sigma$ and $\mathcal{T}$ via schema discovery, taxonomy lookup, and vertex sampling, replacing model guesses with values from the live graph. The chosen handler also retrieves the top $k$ skills from $\mathcal{S}$ by cosine similarity to $q$, conditioning generation on past runs.
\circled{3} \textbf{Plan.} When $\rho = \textsc{plan}$, $q$ is decomposed into a plan $\mathcal{P}$ of parallel and sequential branches with a final merge step.
\circled{4} \textbf{Verify.} A runtime safety wrapper blocks unsafe Gremlin, an inline plan validator checks $\mathcal{P}$, and a post-hoc property verifier audits captured queries against four similarity-edge invariants.
\circled{5} \textbf{Learn.} Every verifier-passing run is distilled into a skill, regardless of route. The library grows monotonically as evaluation proceeds, so SEGRA conditions later questions on a richer experience archive. The full procedure is formalized in Appendix~\ref{app:algorithm}.

\subsection{Direct and Similarity Handlers}
\label{sec:base}

The Direct ($H_{\textsc{dir}}$) and Similarity ($H_{\textsc{sim}}$) handlers both answer questions in a single LLM tool-use loop that emits one Gremlin query against $G$. They differ in what the question targets: $H_{\textsc{dir}}$ filters and aggregates over case properties and the taxonomy (e.g., \emph{``how many VPN cases were resolved last quarter?''}), while $H_{\textsc{sim}}$ traverses derived similarity edges between cases to retrieve solved history (e.g., \emph{``what resolutions worked for cases similar to V123?''}). Both ground generation in $\Sigma$ and $\mathcal{T}$, since the model has no view of the live graph and would otherwise have to guess vertex labels and taxonomy values from training-time knowledge. Both also retrieve relevant skills from $\mathcal{S}$ before generation (§\ref{sec:skills}), giving the model in-context Gremlin patterns that succeeded on similar past questions.


\noindent{\bf Direct ($H_{\textsc{dir}}$).} This handler runs an LLM tool-use loop with five tools.
\roundbox{\textsc{schema-discovery}} retrieves $\Sigma$ from the live graph $G$ and injects it into the system prompt.
\roundbox{\textsc{taxonomy-lookup}} resolves user phrasing to values in the hierarchical taxonomy $\mathcal{T}$, returning matching entries ranked into \emph{strong} matches (the topic appears as a token in the entry name) and \emph{weak} matches (the topic only appears in the description); strong matches are used directly in \texttt{P.within(...)} filters, while weak matches are surfaced with explicit instructions to include them only when the description confirms relevance.
\roundbox{\textsc{vertex-sampling}} returns up to three real vertices of a given label with their property values, enabling the model to verify property formats before composing filters.
\roundbox{\textsc{text-search}} runs \texttt{TextP.containing(\textit{topic})} against the free-text \texttt{content} property of \texttt{reason}, \texttt{symptoms}, and \texttt{rootCause} vertices and reports per-label case counts plus three sample texts, recovering questions whose topics lack taxonomy entries.
\roundbox{\textsc{execute-gremlin}} forwards the LLM-emitted Gremlin to the runtime safety wrapper (§\ref{sec:guards}) and then to $G$, returning results back into the loop.

\noindent{\bf Similarity ($H_{\textsc{sim}}$).} This handler runs the base loop with three additional tools that cover two execution patterns: an \emph{anchored walk} from a referenced case, and a \emph{free-text seeded walk} when no case is referenced. The intent router selects the pattern by checking $q$ for a case-ID, so the handler dispatches to the right tools without runtime disambiguation.

The anchored walk handles questions like \emph{``what is similar to case V123?''}. \roundbox{\textsc{resolve-case}} verifies the referenced case exists and reports which of its content links (to reason, symptoms, root cause) are populated, so the model picks a feasible similarity edge. \roundbox{\textsc{similar-cases}} then executes the canonical bidirectional walk ($\textit{case} \to \textit{entity} \to \texttt{bothE}(\texttt{similar*}) \to \textit{entity} \to \textit{case}$) with a score threshold, avoiding the asymmetric-traversal failure mode where directional \texttt{outE} drops a substantial fraction of peers.

The free-text seeded walk handles questions like \emph{``find cases like a VPN authentication failure''}. \roundbox{\textsc{vector-search-cases}} retrieves cases by free-text similarity in embedding space, providing seeds that the model can optionally chain into \roundbox{\textsc{similar-cases}} to expand. The post-hoc property verifier (§\ref{sec:guards}) audits captured similarity queries from both patterns offline.


\subsection{Plan: Multi-shot Decomposition}
\label{sec:planner}

The Plan handler $H_{\textsc{plan}}$ targets questions that no single Gremlin query can answer, such as comparisons (\emph{``compare resolution rate of VPN vs. Outlook''}), set differences, and top-$N$-then-aggregate queries. It decomposes the question into a structured plan $\mathcal{P}$ of dependent branches, dispatches each branch through the router, and merges branch outputs into a final answer.

\noindent{\bf Operators.} $H_{\textsc{plan}}$ exposes three tools to the LLM.
\roundbox{\textsc{plan-query}} produces $\mathcal{P}$, specifying branches with their dependencies, sub-question texts, target handlers $\rho_b \in \mathcal{R} \setminus \{\textsc{plan}\}$, and a merge strategy; the plan is validated inline (§\ref{sec:guards}) before execution.
\roundbox{\textsc{run-branch}} dispatches each branch recursively through the router into $H_{\rho_b}$; independent branches run in parallel.
\roundbox{\textsc{merge-results}} combines branch outputs to produce the final answer without further graph traversal.

\noindent{\bf Use of retrieved skills.} Like the Direct and Similarity handlers, $H_{\textsc{plan}}$ retrieves skills from $\mathcal{S}$ at the Ground stage, but uses them more heavily: retrieved skills surface past plan structures and per-branch Gremlin templates that the planner adapts to the new question. Multi-shot questions benefit most from this guidance, since past worked examples constrain a much larger decision space — where to split the question, what merge strategy applies, and how each branch should be phrased — than single-query routes face.

\subsection{Verify: Correctness Mechanisms}
\label{sec:guards}

SEGRA enforces correctness through three mechanisms at different stages of the pipeline: a runtime sanitiser and a plan validator guard execution inline, while a post-hoc property verifier audits captured queries offline and gates skill extraction.

\noindent{\bf Runtime safety wrapper.} Every Gremlin query emitted by \roundbox{\textsc{execute-gremlin}} passes through a sanitiser before reaching $G$. It strips markdown fences, blocks mutations, rejects full-graph scans, and auto-injects \texttt{.limit($N$)} on non-aggregating queries. We extend the blocklist with bare similarity-edge scans (e.g., \texttt{g.E().hasLabel('similarReason')}), which are unsafe on graphs with millions of edges per similarity label. Violations return to the tool-use loop as structured errors for correction.

\noindent{\bf Inline plan validator.} Plans from \roundbox{\textsc{plan-query}} are validated before execution: each branch must declare a route hint $\rho_b \in \mathcal{R} \setminus \{\textsc{plan}\}$, the merge strategy must come from a fixed vocabulary, the dependency graph must be acyclic, and recursion through \roundbox{\textsc{run-branch}} is bounded. Invalid plans are rejected, preventing malformed decompositions from reaching $G$.

\noindent{\bf Post-hoc property verifier.} A regex-based offline checker audits four similarity-edge correctness properties on the captured Gremlin history (full definitions in Appendix~\ref{app:verifier}). The verifier walks every captured query regardless of which handler produced it. A run that passes all properties counts as \emph{verifier-pass}, which gates skill extraction (§\ref{sec:skills}).


\subsection{Learn: Skill Library}
\label{sec:skills}

The skill library $\mathcal{S}$ persists experience across runs with an in-memory cosine-similarity index over question embeddings. All three handlers ($H_{\textsc{dir}}$, $H_{\textsc{sim}}$, $H_{\textsc{plan}}$) consult $\mathcal{S}$ during the Ground stage and contribute to it after a verifier-passing run.

\noindent{\bf Skill schema.} A skill records the question template, the route that answered it, and the captured Gremlin templates with entity-specific tokens abstracted; multi-shot skills additionally record the plan structure and merge strategy.

\noindent{\bf Retrieval.} For each question, the library returns the top $k$ skills by cosine similarity to the question embedding. $H_{\textsc{dir}}$ and $H_{\textsc{sim}}$ see worked Gremlin templates as in-context examples; $H_{\textsc{plan}}$ additionally sees plan structures it can adapt.


\noindent{\bf Online accumulation.} On the shuffled evaluation set, SEGRA processes questions sequentially: each handler retrieves from $\mathcal{S}$, generates, and writes back a new skill when its run is verifier-pass and judged successful. Each question therefore retrieves from a strictly larger library, accumulating experience across the evaluation sequence rather than across separate training and test phases. We isolate this effect with a baseline that runs the same questions in the same order with $\mathcal{S}$ disabled (§\ref{sec:results}).

\section{Experimental Setup}\label{sec:setup}

We evaluate SEGRA on an enterprise IT support QA graph using a benchmark of 156 natural-language questions that mirror the analytical and diagnostic queries operators issue in production. The questions are constructed by the authors from recurring enterprise IT support information needs, spanning direct lookup, taxonomy-based aggregation, similarity-driven retrieval, and multi-step diagnostic analysis. Our primary deployment-oriented backbone is a commercially hosted instruction-following LLM in the \texttt{Medium} tier, selected to reflect a strong cost–quality tradeoff for practical deployment. To assess whether SEGRA's gains depend on backbone capability, we also evaluate \texttt{Small} and \texttt{Large} tier backbones under the same protocol. We anonymize the backbone identities to keep the evaluation focused on the SEGRA framework rather than on benchmarking or ranking specific commercial models.\footnote{The underlying commercial backbone identities are omitted due to business confidentiality constraints. The tier assignments are fixed across all experiments, and all systems are evaluated under the same protocol.}
Full details of the graph database and question set are in \cref{app:setup}. 

\subsection{Systems Comparison}\label{sec:systems}

We compare three systems that together stage the contribution of SEGRA into a backbone-only baseline, the architecture without skill memory, and the architecture with skill memory. All three share the same generator LLM, graph access, evaluation order, and validation protocol, and differ only in which mechanisms they expose. 
\circled{1} \textbf{CoT.} A single LLM call asked to emit a Gremlin query with chain-of-thought reasoning. No live schema, no tools, no skill library. Establishes the floor: how well a generic LLM can guess at an unseen schema.
\circled{2} \textbf{Few-shot prompting CoT.} A single LLM call with the live schema $\Sigma$ in the prompt and five worked (question, Gremlin) examples, but no tools, no execution feedback, and no skill library. Tests whether richer prompting alone, without an agent loop, can close the gap to SEGRA.
\circled{3} \textbf{SEGRA w/o skills.} Full SEGRA architecture (router, planner, handlers, tool surface, post-hoc verifier) with the skill library $\mathcal{S}$ disabled: retrieval returns the empty set, and runs do not write skills back. Every question is dispatched through the same handler graph as SEGRA w/ skills, so this isolates the contribution of $\mathcal{S}$ from the contribution of the architecture itself.
\circled{4} \textbf{SEGRA w/ skills.} The complete system. The router classifies intent, retrieved skills condition both the planner outer loop and every dispatched handler sub-loop, and verified runs write skills back into $\mathcal{S}$ via Alg.~\ref{alg:segra-library}. Matches §\ref{sec:method} and is the headline system in §\ref{sec:results}.


\subsection{Evaluation Metrics}\label{sec:metrics}


\noindent{\bf LLM-as-judge.} A judge LLM scores each run on five orthogonal dimensions -- \emph{intent fidelity}, \emph{execution validity}, \emph{grounding}, \emph{completeness}, and \emph{usefulness} -- each on a 0--4 scale, with the aggregate \texttt{overall\_score} taken as the floor of the mean. The judge sees the question, the generated Gremlin, the executed result, and the answer, but not which system produced the run. 
We use OpenAI GPT-5.4 with medium reasoning as the automatic judge. Since the generator backbones (§\ref{sec:setup}) come from a different model provider, this choice avoids the risk of self-preference in evaluation.
Full per-dimension definitions and score anchors are in Appendix~\ref{app:judge}.

\noindent{\bf Cost and efficiency.} For each run we log the tool-use \emph{turns} taken by the generator (one turn per LLM call, planner outer loop plus all sub-loops) and the \emph{token totals}: input tokens, output tokens, cache-read input tokens, and cache-creation input tokens. We report mean tokens and mean turns per question for each system, broken down by reasoning shape, so the cost of routing, decomposition, and skill retrieval can be read alongside their accuracy contribution. Multi-shot questions accumulate tokens and turns across the planner outer loop and every branch sub-run; we report the per-question total to keep comparisons end-to-end.

\section{Results}
\label{sec:results}

This section compares SEGRA with the baselines in §\ref{sec:systems}. We report results across \texttt{Small}, \texttt{Medium}, \texttt{Large} generator backbones to test robustness across model tiers.

\subsection{Main Results}
\label{sec:main}

\cref{tab:main-results} reports mean judge score, mean LLM calls, and mean dollar cost per question. SEGRA changes the performance regime: backbone-only CoT performs poorly on all model tiers, and adding the schema with five worked examples (Few-shot CoT) raises score by $0.16$--$0.29$ points but is still far short of the SEGRA architecture. Even \texttt{Large} with Few-shot CoT scores $0.47$, below \texttt{Small} running SEGRA without skills ($1.48$). This shows that neither model scale nor richer prompting alone solves enterprise text-to-Gremlin QA; the system needs schema-grounded planning, tool use, execution feedback, and verification.

The agent architecture accounts for the largest quality gain. Moving from Few-shot CoT to SEGRA without skills improves mean score by $+1.07$ on \texttt{Small}, $+1.44$ on \texttt{Medium}, and $+1.56$ on \texttt{Large}. On our primary \texttt{Medium} backbone, SEGRA without skills lifts the score from $0.50$ to $1.94$, bringing the system close to the judge pass threshold. With skills enabled, \texttt{Medium} reaches $2.03$, crossing the threshold while reducing the average number of LLM calls from $11.3$ to $9.0$. 
Few-shot CoT's failure mode is concentrated on multi-shot questions: with no opportunity to decompose, retry, or self-correct.
SEGRA's planner and handler loops convert these into branch-and-merge plans with per-branch execution feedback, which is what the architecture lift in the table reflects.

The skill library mainly improves efficiency rather than replacing the architecture. Across all three backbones, adding $\mathcal{S}$ preserves or slightly improves quality while reducing both calls and cost: calls drop by $21\%$ on \texttt{Small}, $20\%$ on \texttt{Medium}, and $26\%$ on \texttt{Large}, with corresponding cost reductions of $19\%$, $18\%$, and $28\%$. Thus, SEGRA's two components play complementary roles: {\bf the agent architecture provides the quality lift, and the skill library recovers a substantial fraction of the tool-use overhead by reusing verified query patterns.}

\subsection{Model-Tier Trade-offs}

\begin{figure}[t]
\centering
\includegraphics[width=\linewidth]{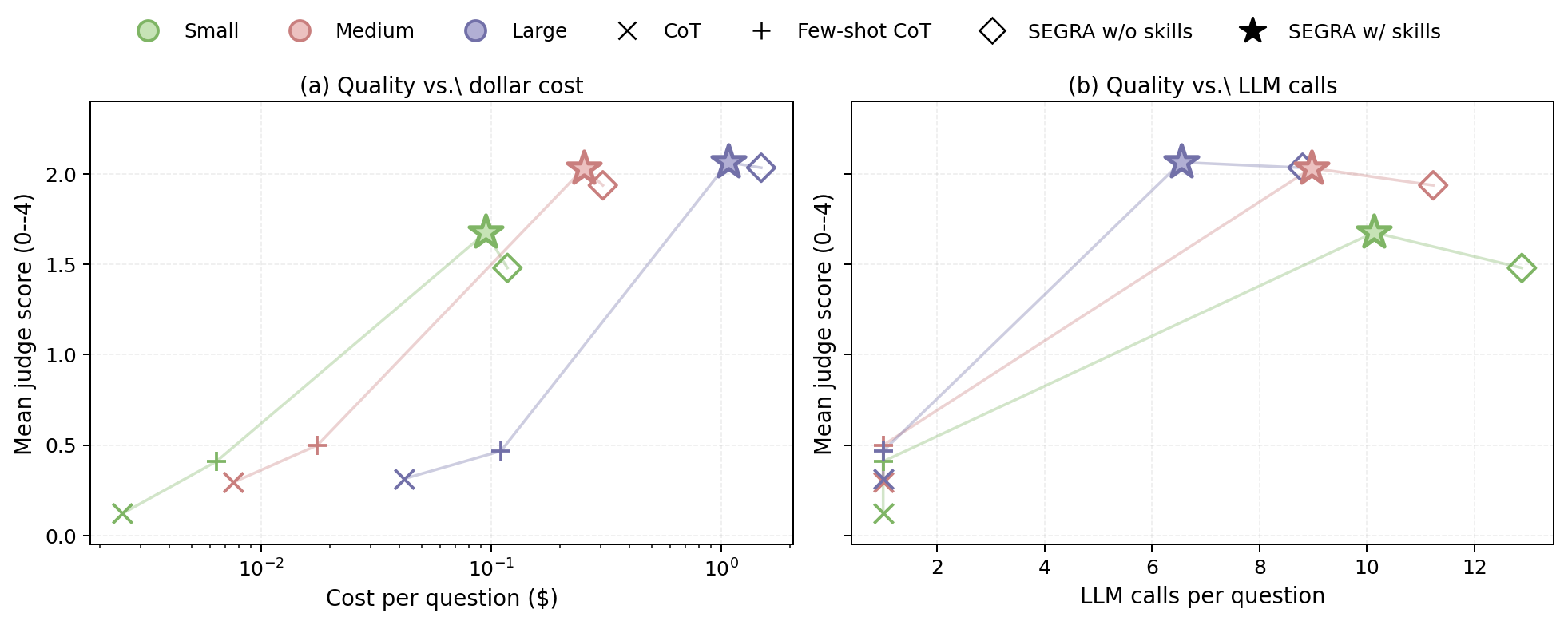}
\caption{Quality--cost and quality--calls trade-offs. SEGRA architecture lifts every backbone to better score and lower cost and fewer calls.
}
\label{fig:cost}
\vspace{-0.5cm}
\end{figure}

\cref{fig:cost} compares quality against runtime cost and LLM calls across model tiers. SEGRA improves the quality--cost trade-off substantially: \texttt{Medium} with SEGRA and skills reaches the same mean score as \texttt{Large} with SEGRA without skills ($2.03$), while costing roughly 1/6 as much. It is also within $0.03$ points of the best-scoring configuration, \texttt{Large} with skills, at less than 1/4 of the cost. These results show that system design can partially substitute for model scale. The SEGRA architecture provides the main quality gain, while the skill library improves efficiency by reducing calls and cost without degrading quality. In practical terms, SEGRA makes a mid-tier backbone competitive with a substantially larger model, and skills improve the efficiency of every backbone.

\begin{table}[t]
\centering
\small
\begin{adjustbox}{width=\linewidth}
\begin{tabular}{l rrr rrr rrr}
\toprule
& \multicolumn{3}{c}{\texttt{Small}} & \multicolumn{3}{c}{\texttt{Medium}} & \multicolumn{3}{c}{\texttt{Large}} \\
\cmidrule(lr){2-4} \cmidrule(lr){5-7} \cmidrule(lr){8-10}
System & Score & Calls & \$/q & Score & Calls & \$/q & Score & Calls & \$/q \\
\midrule
CoT             & 0.12 & \phantom{0}1.0 & 0.003 & 0.29 & \phantom{0}1.0 & 0.008 & 0.31 & \phantom{0}1.0 & 0.042 \\
Few-shot CoT    & 0.41 & \phantom{0}1.0 & 0.006 & 0.50 & \phantom{0}1.0 & 0.018 & 0.47 & \phantom{0}1.0 & 0.111 \\
SEGRA w/o $\mathcal{S}$                & 1.48 & 12.9 & 0.118 & 1.94 & 11.3 & 0.307 & 2.03 & \phantom{0}8.8 & 1.498 \\
\rowcolor[HTML]{EFEFEF}
\textbf{SEGRA w/ $\mathcal{S}$} & \textbf{1.67} & \textbf{10.2} & \textbf{0.095} & \textbf{2.03} & \phantom{0}\textbf{9.0} & \textbf{0.253} & \textbf{2.06} & \phantom{0}\textbf{6.5} & \textbf{1.074} \\
\bottomrule
\end{tabular}
\end{adjustbox}
\caption{Mean judge score, mean LLM calls per question, and mean cost per question by system and backbone. SEGRA improves quality substantially, and the skill library further reduces cost/calls. 
}
\label{tab:main-results}
\end{table}

\subsection{Human Evaluation}
\label{sec:human}

We manually evaluate SEGRA w/ $\mathcal{S}$ with the \texttt{Medium} backbone on all 156 benchmark questions. Two annotators independently score each output using the same 0--4 rubric as the LLM judge. Results are reported by handler: direct querying ($H_{\textsc{DIR}}$, $n=36$), similarity traversal ($H_{\textsc{SIM}}$, $n=20$), and planner-based multi-shot reasoning ($H_{\textsc{PLAN}}$, $n=100$). Annotators see the question, generated Gremlin, execution result, and final answer, but not the judge score; LLM-generated Gremlin explanations are used only as annotation aids. Detailed inter-annotator results are included in \cref{app:human-judge}.

\begin{figure}[t!]
\centering
\includegraphics[width=0.7\linewidth]{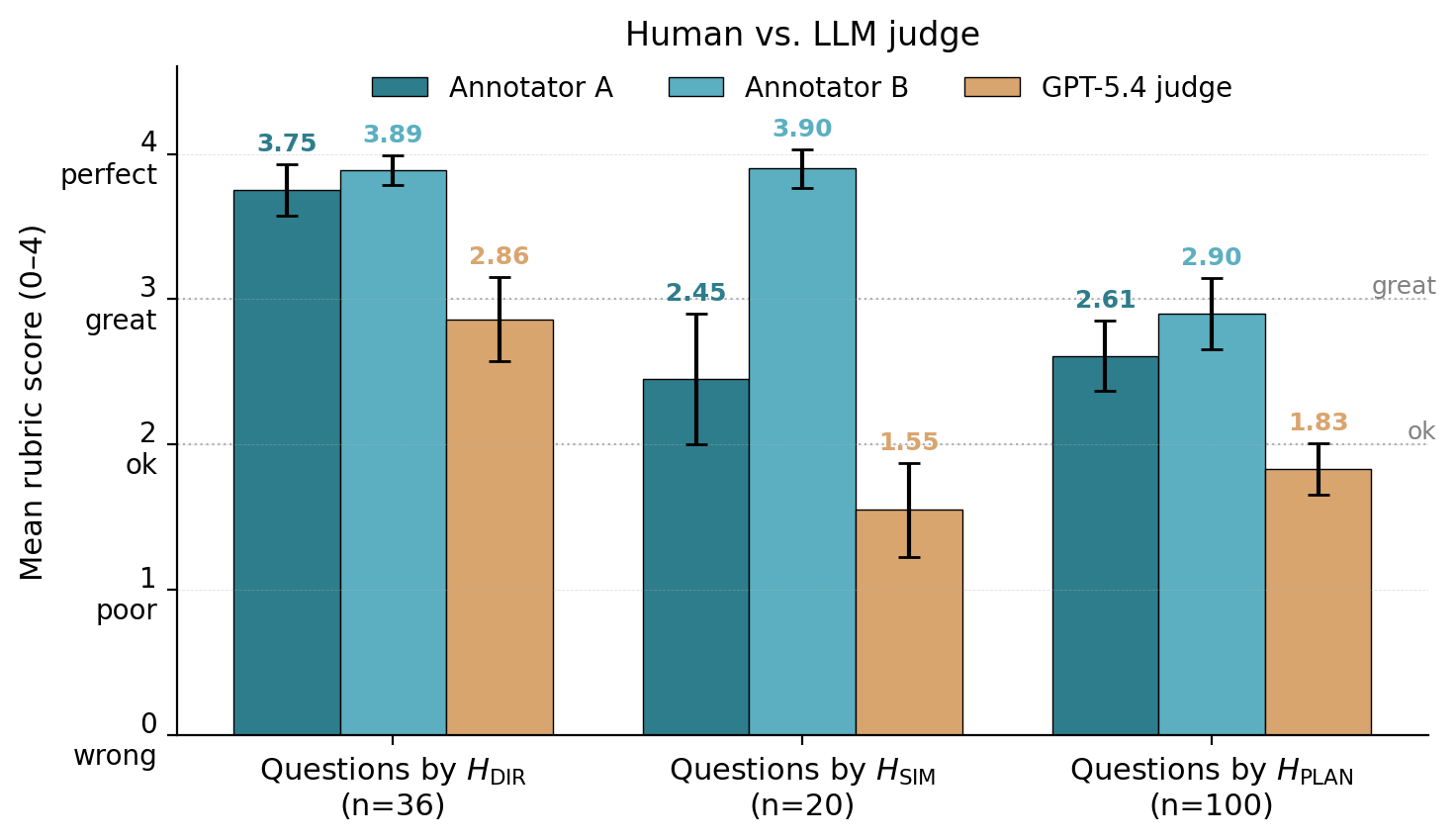}
\caption{Human and LLM judge mean scores for SEGRA w/ $\mathcal{S}$, stratified by handler. Error bars show 95\% confidence intervals.}
\label{fig}
\vspace{-0.5cm}
\end{figure}

Figure~\ref{fig} shows that both annotators rate SEGRA higher than the automatic judge across all handlers. Averaged human scores exceed judge scores by $0.96$ on $H_{\textsc{DIR}}$, $1.62$ on $H_{\textsc{SIM}}$, and $0.92$ on $H_{\textsc{PLAN}}$. The disagreement is one-sided: among the 35 questions where the averaged human score differs from the judge by at least two points, the judge is lower in every case. $H_{\textsc{DIR}}$ is the strongest handler, with over 80\% of direct-query outputs rated perfect by both annotators and no poor or wrong cases. Overall, the judge provides a conservative scalable metric, especially for similarity and planner outputs; we therefore interpret judge scores as lower-bound estimates of absolute quality and complement them with a strict count-accuracy analysis in \cref{app:exec-match}.

\section{Conclusion}

We presented SEGRA, an experience-guided framework for Gremlin-based question answering over enterprise IT support graphs. SEGRA combines schema- and taxonomy-grounded query generation, similarity-aware traversal handling, planner-based decomposition, execution-aware verification, and a skill library that reuses verified query patterns across questions. Experiments show that SEGRA substantially improves over backbone-only chain-of-thought prompting, while the skill library reduces LLM calls and cost without degrading answer quality. Human evaluation further suggests that automatic judge scores are conservative, especially for planner-based multi-step outputs. Overall, SEGRA shows that structured agent design and reusable execution experience can make natural-language interfaces to property graphs more accurate, efficient, and deployable.
The agent design (intent routing, multi-shot decomposition and merge, execution-aware verification, and cross-query skill reuse) is independent of Gremlin and
of IT support. The same design therefore transfers to other graph query languages such as Cypher and to other property-graph domains.

\section*{Limitations}

SEGRA is evaluated on a single enterprise IT support graph, so broader application on other property graphs remains to be validated. The system also assumes access to schema tools, execution feedback, and verifier rules. In longer deployments, the skill library may require maintenance under schema drift. Finally, human evaluation covers only the primary configuration and we use the automatic judge for the full system-by-backbone comparison, leaving broader human evaluation for future work.





\bibliography{references}

@article{huang2022dgraph,
  title={Dgraph: A large-scale financial dataset for graph anomaly detection},
  author={Huang, Xuanwen and Yang, Yang and Wang, Yang and Wang, Chunping and Zhang, Zhisheng and Xu, Jiarong and Chen, Lei and Vazirgiannis, Michalis},
  journal={Advances in Neural Information Processing Systems},
  volume={35},
  pages={22765--22777},
  year={2022}
}

@article{guo2020survey,
title={A survey on knowledge graph-based recommender systems},
author={Guo, Qingyu and Zhuang, Fuzhen and Qin, Chuan and Zhu, Hengshu and Xie, Xing and Xiong, Hui and He, Qing},
journal={IEEE Transactions on Knowledge and Data Engineering},
volume={34},
number={8},
pages={3549--3568},
year={2020},
publisher={IEEE}
}

@article{besta2023demystifying,
title={Demystifying graph databases: Analysis and taxonomy of data organization, system designs, and graph queries},
author={Besta, Maciej and Gerstenberger, Robert and Peter, Emanuel and Fischer, Marc and Podstawski, Micha{\l} and Barthels, Claude and Alonso, Gustavo and Hoefler, Torsten},
journal={ACM Computing Surveys},
volume={56},
number={2},
pages={1--40},
year={2023},
publisher={ACM New York, NY}
}

@article{angles2017foundations,
title={Foundations of modern query languages for graph databases},
author={Angles, Renzo and Arenas, Marcelo and Barcel{\'o}, Pablo and Hogan, Aidan and Reutter, Juan and Vrgo{\v{c}}, Domagoj},
journal={ACM Computing Surveys (CSUR)},
volume={50},
number={5},
pages={1--40},
year={2017},
publisher={ACM New York, NY, USA}
}

@inproceedings{rodriguez2015gremlin,
title={The gremlin graph traversal machine and language (invited talk)},
author={Rodriguez, Marko A},
booktitle={Proceedings of the 15th symposium on database programming languages},
pages={1--10},
year={2015}
}

@inproceedings{francis2018cypher,
  title={Cypher: An evolving query language for property graphs},
  author={Francis, Nadime and Green, Alastair and Guagliardo, Paolo and Libkin, Leonid and Lindaaker, Tobias and Marsault, Victor and Plantikow, Stefan and Rydberg, Mats and Selmer, Petra and Taylor, Andr{\'e}s},
  booktitle={Proceedings of the 2018 international conference on management of data},
  pages={1433--1445},
  year={2018}
}

@article{chen2024beaver,
title={BEAVER: an enterprise benchmark for text-to-sql},
author={Chen, Peter Baile and Yang, Devin and Li, Weiyue and Wenz, Fabian and Zhang, Yi and Tatbul, Nesime and Cafarella, Michael and Demiralp, {\c{C}}a{\u{g}}atay and Stonebraker, Michael},
journal={arXiv preprint arXiv:2409.02038},
year={2024}
}

@article{pourreza2023din,
title={Din-sql: Decomposed in-context learning of text-to-sql with self-correction},
author={Pourreza, Mohammadreza and Rafiei, Davood},
journal={Advances in neural information processing systems},
volume={36},
pages={36339--36348},
year={2023}
}

@inproceedings{wang2025mac,
title={Mac-sql: A multi-agent collaborative framework for text-to-sql},
author={Wang, Bing and Ren, Changyu and Yang, Jian and Liang, Xinnian and Bai, Jiaqi and Chai, Linzheng and Yan, Zhao and Zhang, Qian-Wen and Yin, Di and Sun, Xing and others},
booktitle={Proceedings of the 31st International Conference on Computational Linguistics},
pages={540--557},
year={2025}
}

@article{lyu2026text2gql,
title={Text2GQL-Bench: A Text to Graph Query Language Benchmark [Experiment, Analysis \& Benchmark]},
author={Lyu, Songlin and Ban, Lujie and Wu, Zihang and Luo, Tianqi and Liu, Jirong and Ma, Chenhao and Luo, Yuyu and Tang, Nan and Qi, Shipeng and Lin, Heng and others},
journal={arXiv preprint arXiv:2602.11745},
year={2026}
}

@inproceedings{ozsoy2025text2cypher,
title={Text2cypher: Bridging natural language and graph databases},
author={Ozsoy, Makbule Gulcin and Messallem, Leila and Besga, Jon and Minneci, Gianandrea},
booktitle={Proceedings of the Workshop on Generative AI and Knowledge Graphs (GenAIK)},
pages={100--108},
year={2025}
}

@inproceedings{yu2018spider,
  title={Spider: A large-scale human-labeled dataset for complex and cross-domain semantic parsing and text-to-sql task},
  author={Yu, Tao and Zhang, Rui and Yang, Kai and Yasunaga, Michihiro and Wang, Dongxu and Li, Zifan and Ma, James and Li, Irene and Yao, Qingning and Roman, Shanelle and others},
  booktitle={Proceedings of the 2018 conference on empirical methods in natural language processing},
  pages={3911--3921},
  year={2018}
}

@article{li2023can,
  title={Can llm already serve as a database interface? a big bench for large-scale database grounded text-to-sqls},
  author={Li, Jinyang and Hui, Binyuan and Qu, Ge and Yang, Jiaxi and Li, Binhua and Li, Bowen and Wang, Bailin and Qin, Bowen and Geng, Ruiying and Huo, Nan and others},
  journal={Advances in Neural Information Processing Systems},
  volume={36},
  pages={42330--42357},
  year={2023}
}

@inproceedings{sun2024think,
  title={Think-on-graph: Deep and responsible reasoning of large language model on knowledge graph},
  author={Sun, Jiashuo and Xu, Chengjin and Tang, Lumingyuan and Wang, Saizhuo and Lin, Chen and Gong, Yeyun and Ni, Lionel and Shum, Heung-Yeung and Guo, Jian},
  booktitle={International Conference on Learning Representations},
  volume={2024},
  pages={3868--3898},
  year={2024}
}

@inproceedings{luo2024chatkbqa,
  title={Chatkbqa: A generate-then-retrieve framework for knowledge base question answering with fine-tuned large language models},
  author={Luo, Haoran and Haihong, E and Tang, Zichen and Peng, Shiyao and Guo, Yikai and Zhang, Wentai and Ma, Chenghao and Dong, Guanting and Song, Meina and Lin, Wei and others},
  booktitle={Findings of the association for computational linguistics: ACL 2024},
  pages={2039--2056},
  year={2024}
}

@article{wang2023voyager,
  title={Voyager: An open-ended embodied agent with large language models},
  author={Wang, Guanzhi and Xie, Yuqi and Jiang, Yunfan and Mandlekar, Ajay and Xiao, Chaowei and Zhu, Yuke and Fan, Linxi and Anandkumar, Anima},
  journal={arXiv preprint arXiv:2305.16291},
  year={2023}
}

@inproceedings{zhang2025aflow,
  title={Aflow: Automating agentic workflow generation},
  author={Zhang, Jiayi and Xiang, Jinyu and Yu, Zhaoyang and Teng, Fengwei and Chen, Xionghui and Chen, Jiaqi and Zhuge, Mingchen and Cheng, Xin and Hong, Sirui and Wang, Jinlin and others},
  booktitle={International Conference on Learning Representations},
  volume={2025},
  pages={34040--34077},
  year={2025}
}

@article{shinn2023reflexion,
  title={Reflexion: Language agents with verbal reinforcement learning},
  author={Shinn, Noah and Cassano, Federico and Gopinath, Ashwin and Narasimhan, Karthik and Yao, Shunyu},
  journal={Advances in neural information processing systems},
  volume={36},
  pages={8634--8652},
  year={2023}
}

@article{zhang2025darwin,
  title={Darwin godel machine: Open-ended evolution of self-improving agents},
  author={Zhang, Jenny and Hu, Shengran and Lu, Cong and Lange, Robert and Clune, Jeff},
  journal={arXiv preprint arXiv:2505.22954},
  year={2025}
}

@article{kapoor2026dqa,
  title={DQA: Diagnostic Question Answering for IT Support},
  author={Kapoor, Vishaal and Dundua, Mariam and Ahuja, Sarthak and Kordjazi, Neda and Yortucboylu, Evren and Padala, Vaibhavi and Ho, Derek and Whitted, Jennifer and Steinert, Rebecca},
  journal={arXiv preprint arXiv:2604.05350},
  year={2026}
}

@article{ahuja2026vigil,
  title={VIGIL: Towards Edge-Extended Agentic AI for Enterprise IT Support},
  author={Ahuja, Sarthak and Kordjazi, Neda and Yortucboylu, Evren and Kapoor, Vishaal and Dundua, Mariam and Li, Yiming and Ho, Derek and Padala, Vaibhavi and Whitted, Jennifer and Steinert, Rebecca},
  journal={arXiv preprint arXiv:2603.16110},
  year={2026}
}

@inproceedings{ahmed2023recommending,
  title={Recommending root-cause and mitigation steps for cloud incidents using large language models},
  author={Ahmed, Toufique and Ghosh, Supriyo and Bansal, Chetan and Zimmermann, Thomas and Zhang, Xuchao and Rajmohan, Saravan},
  booktitle={2023 IEEE/ACM 45th International Conference on Software Engineering (ICSE)},
  pages={1737--1749},
  year={2023},
  organization={IEEE}
}

@inproceedings{roy2024exploring,
  title={Exploring llm-based agents for root cause analysis},
  author={Roy, Devjeet and Zhang, Xuchao and Bhave, Rashi and Bansal, Chetan and Las-Casas, Pedro and Fonseca, Rodrigo and Rajmohan, Saravan},
  booktitle={Companion proceedings of the 32nd ACM international conference on the foundations of software engineering},
  pages={208--219},
  year={2024}
}

@article{chen2025aiopslab,
  title={Aiopslab: A holistic framework to evaluate ai agents for enabling autonomous clouds},
  author={Chen, Yinfang and Shetty, Manish and Somashekar, Gagan and Ma, Minghua and Simmhan, Yogesh and Mace, Jonathan and Bansal, Chetan and Wang, Rujia and Rajmohan, Saravan},
  journal={Proceedings of Machine Learning and Systems},
  volume={7},
  year={2025}
}

@inproceedings{yun2025eicopilot,
  title={EICopilot: Search and Explore Enterprise Information over Large-scale Knowledge Graphs with LLM-driven Agents},
  author={Yun, Yuhui and Ye, Huilong and Deng, Jingfeng and Li, Ruojia and Li, Xinru and Li, Li and Xiong, Haoyi},
  booktitle={2025 IEEE International Conference on Big Data (BigData)},
  pages={2689--2696},
  year={2025},
  organization={IEEE}
}

\appendix

\section{Strict Count-Accuracy Evaluation}
\label{app:exec-match}

To further study answer quality under deterministic ground truth, we construct a scalar-count analytic subset from the 156-question evaluation benchmark. We include only questions whose gold solution can be represented as a single Gremlin query returning a scalar count. For each included question, a human annotator manually writes the gold Gremlin query, executes it on the graph, and uses the returned count as the reference answer. This filtering excludes display-style questions, list-valued answers, schema-introspection questions, free-form explanations, and multi-shot questions whose gold solution requires decomposition. The resulting subset contains $n=60$ questions. It is not randomly sampled; it is the eligible subset for which human-written query ground truth and scalar-count comparison are well defined.

For each question $q_i$, let $g_i$ be the human-written gold Gremlin query and let $\widehat{g}_i$ be the query produced by SEGRA. 
We report two diagnostics. \textbf{Query match} measures whether the generated query has the same structure as the human-written gold query after a normalizer $C(\cdot)$ removes known meaning-preserving differences, such as formatting, equivalent interval predicates, symbolic time-window literals, contact--case anchoring variants, and result-preserving deduplication. For example, a generated query using \texttt{P.between(a,b)} and a gold query using \texttt{P.gte(a).and(P.lt(b))} are treated as matching if the two traversals are otherwise identical. Similarly, harmless differences in whitespace, quote style, or redundant label checks after a fixed vertex anchor are ignored. 
The metric is exact equality after normalization:
\[
\mathrm{Acc}_{\mathrm{query}}
=
\frac{1}{n}\sum_{i=1}^{n}
\mathds{1}\bigl\{C(\hat g_i)=C(g_i)\bigr\}.
\]
\textbf{Execution match} measures whether the generated query returns the same scalar count as the human-written gold query. 
Let $y_i=\mathrm{Exec}(g_i)$ and $\widehat{y}_i=\mathrm{Exec}(\widehat{g}_i)$. 
We count an execution as correct when
\[
\mathrm{Acc}_{\mathrm{exec}}
=
\frac{1}{n}\sum_{i=1}^{n}
\mathds{1}\!\Bigl\{\frac{
\bigl|\hat y_i-y_i\bigr|}{\max(|y_i|,1)}\le 0.02
\Bigr\}.
\]
We report the average query-match and execution-match accuracy over the 60-question subset. Query match evaluates agreement with the manually written Gremlin structure; execution match evaluates agreement with the human-established scalar answer.

\paragraph{Results.}
Table~\ref{tab:exec-match} reports both metrics for SEGRA w/ $\mathcal{S}$ across the three backbones. \texttt{Medium} achieves the highest execution match ($54\%$), while \texttt{Large} achieves the highest query match ($48\%$). This shows that structural agreement and answer agreement are related but not identical: a model may generate a different traversal while still computing the correct count.

\begin{table}[h!]
\centering
\small
\begin{tabular}{lcc}
\toprule
Backbone & $\mathrm{Acc}_{\mathrm{query}}$ & $\mathrm{Acc}_{\mathrm{exec}}$ \\
\midrule
\texttt{Small}  & $35\%$ & $41\%$ \\
\texttt{Medium} & $40\%$ & $\mathbf{54\%}$ \\
\texttt{Large}   & $\mathbf{48\%}$ & $44\%$ \\
\bottomrule
\end{tabular}
\caption{Query match and execution match on the $n=60$ scalar-count analytic subset for SEGRA w/ $\mathcal{S}$. 
}
\label{tab:exec-match}
\end{table}

This subset analysis complements the judge-based and human evaluations by isolating cases where exact scalar-count comparison against human-written Gremlin ground truth is possible. We treat these metrics as diagnostics rather than headline quality measures because they do not capture partial correctness, alternative valid decompositions, or all semantically equivalent Gremlin formulations.

We also reflect the results in \cref{tab:exec-match} that the absolute correctness is relatively limited compared to stronger judge and human ratings. We identify three main sources of disagreement: (1) Taxonomy ambiguity. The enterprise taxonomy is large, noisy, and hierarchical. A term in a question may plausibly refer to an issue category, subcategory, issue detail, device attribute, or textual mention. Queries based on different reasonable taxonomy interpretations can preserve the user’s high-level intent but select different populations and return different counts. (2) Temporal and graph-snapshot mismatch. The enterprise graph grows continuously. To make evaluation reproducible, SEGRA is instructed to constrain queries to a fixed time window. Some reference queries omit this window, apply it at a different point in the traversal, or contain answers generated from an earlier graph state. Questions whose temporal scope overlaps or falls outside the fixed window can consequently produce semantically reasonable but numerically different answers. (3) Genuine query-generation errors. Some generated queries omit required taxonomy, text, status, or temporal constraints; traverse the wrong edge or direction; or use incorrect counting, grouping, or deduplication semantics. These are genuine failures rather than alternative interpretations. Thus, the gap should not be interpreted solely as a contradiction between the judge and execution evaluation. The judge assesses whether an answer is semantically appropriate and grounded in its executed result, whereas execution correctness tests agreement with one reference interpretation within a numerical tolerance.

\section{Full Algorithm}
\label{app:algorithm}

This appendix gives the complete SEGRA evaluation procedure, factored into three algorithms: the outer question-stream loop (Alg.~\ref{alg:segra-main}), the multi-shot decomposition that runs when the router selects the plan handler (Alg.~\ref{alg:segra-planner}), and the post-run skill update that gates the library on the verifier (Alg.~\ref{alg:segra-library}). All three operate on the live property graph $G$ with schema $\Sigma$, the domain taxonomy $\mathcal{T}$, the routing function $R$, and the persistent skill library $\mathcal{S}$.

\paragraph{Outer loop (Alg.~\ref{alg:segra-main}).} Questions arrive one at a time in a fixed shuffled order, mirroring how an analyst would issue them in production. For each question $q$, SEGRA routes via $R$, retrieves the top $k$ skills from $\mathcal{S}$ by cosine similarity to $q$, and dispatches to the appropriate handler. Direct and Similarity questions run in a single tool-use loop ($H_\rho$); plan questions invoke Alg.~\ref{alg:segra-planner}. After the run, $\textsc{UpdateLibrary}$ inspects the captured run record and writes a skill back to $\mathcal{S}$ if the verifier accepts it. Each subsequent question therefore retrieves from a strictly larger library; the no-memory baseline disables the write step and feeds the empty $\mathcal{S}_0$ to every question.

\paragraph{Multi-shot decomposition (Alg.~\ref{alg:segra-planner}).} The plan handler first emits a plan tree $\mathcal{P}$ with branches, dependencies, target handlers, and a merge strategy, conditioned on the retrieved skills. The plan validator (§\ref{sec:guards}) checks $\mathcal{P}$ before any branch executes; on failure, it returns the specific validation error (e.g., a duplicate branch id, an unknown route hint, a cycle in the dependency graph) to the planner LLM as the tool result. The LLM may correct the plan and re-invoke \roundbox{\textsc{plan-query}} on a subsequent turn, subject to the planner's overall turn budget (14 turns shared across plan, dispatch, and merge). Once a valid plan is accepted, the planner dispatches each branch in $\mathcal{P}$ back through the router into $H_{\rho_b}$ with $\rho_b \neq \textsc{plan}$, so a branch is itself a single-handler tool-use loop. In our implementation, branches are dispatched in topological order of their declared dependencies, with independent branches running in parallel. After all branches return, \roundbox{\textsc{merge-results}} produces the final answer from branch outputs without further graph traversal. The captured Gremlin history of every branch is collected into the run record for the verifier.

\paragraph{Library update (Alg.~\ref{alg:segra-library}).} The post-hoc property verifier $V_{\text{prop}}$ (Appendix~\ref{app:verifier}) runs over every captured Gremlin query in the run record and reports a verifier-pass flag. A skill is written to $\mathcal{S}$ only when the run is both verifier-pass and judged successful by the LLM judge with score $\geq 2$, ensuring that low-quality or structurally invalid runs cannot pollute the library. The skill schema is the same regardless of route (question template, route label, abstracted Gremlin templates), with multi-shot skills additionally recording the plan structure and merge strategy.

\begin{algorithm}[t!]
\caption{SEGRA: question evaluation with skill accumulation.}
\label{alg:segra-main}
\begin{algorithmic}[1]
\Require Question stream $Q$; initial library $\mathcal{S}_0$; router $R$; live schema $\Sigma$; taxonomy $\mathcal{T}$; retrieval budget $k$.
\State $\mathcal{S} \gets \mathcal{S}_0$
\For{$q \in Q$ in fixed order}
  \State $\rho \gets R(q)$ \Statex \graycomment{intent classification: $\rho \in \{\textsc{dir}, \textsc{sim}, \textsc{plan}\}$}
  \State $\mathcal{K}_q \gets \textsc{Retrieve}(\mathcal{S}, q, k)$
  \Statex \graycomment{top-$k$ skills by cosine similarity}
  \If{$\rho = \textsc{plan}$}
    \State \textit{ans}, \textit{record} $\gets Alg.~\ref{alg:segra-planner}(q, \Sigma, \mathcal{T}, \mathcal{K}_q)$
  \Else
    \State \textit{ans}, \textit{record} $\gets H_\rho(q, \Sigma, \mathcal{T}, \mathcal{K}_q)$ 
    \Statex \graycomment{single-handler tool-use loop}
  \EndIf
  \State $\mathcal{S} \gets Alg.~\ref{alg:segra-library}(\mathcal{S}, \textit{record})$ 
\EndFor
\State \textbf{return} answers and final $\mathcal{S}$
\end{algorithmic}
\end{algorithm}

\begin{algorithm}[t!]
\caption{$\textsc{Plan-Run-Merge}$: multi-shot decomposition invoked from Alg.~\ref{alg:segra-main} when $\rho = \textsc{plan}$.}
\label{alg:segra-planner}
\begin{algorithmic}[1]
\Require $q$, $\Sigma$, $\mathcal{T}$, retrieved skills $\mathcal{K}_q$.
\State $\mathcal{P} \gets \textsc{plan-query}(q, \Sigma, \mathcal{T}, \mathcal{K}_q)$ 
\Statex \graycomment{plan tree with branches and merge $m$}
\If{plan validator rejects $\mathcal{P}$}
  \State \textbf{return} validation error as tool result
\EndIf
\For{branch $b$ in $\mathcal{P}$}
  \State $r_b \gets H_{\rho_b}(q_b, \Sigma, \mathcal{T}, \mathcal{K}_q)$ 
  \Statex \graycomment{recursive dispatch via $R$, $\rho_b \neq \textsc{plan}$}
\EndFor
\State \textit{ans} $\gets \textsc{merge-results}(\{r_b\}_b, m)$ 
\Statex \graycomment{LLM-only step, no graph traversal}
\State \textbf{return} \textit{ans}, run record (plan, branch outputs, Gremlin history)
\end{algorithmic}
\end{algorithm}

\begin{algorithm}[t!]
\caption{$\textsc{UpdateLibrary}$: post-run skill extraction gated by the verifier and the judge.}
\label{alg:segra-library}
\begin{algorithmic}[1]
\Require Library $\mathcal{S}$; run record from Alg.~\ref{alg:segra-main} (Gremlin history, plan, branch outputs, final answer).
\State $\textit{pass} \gets V_{\text{prop}}(\text{record.gremlin\_history})$ 
\Statex \graycomment{Appendix~\ref{app:verifier}}
\State $\textit{score} \gets \textsc{Judge}(\text{record})$ 
\Statex \graycomment{LLM judge, score $\in \{0, 1, 2, 3, 4\}$}
\If{$\textit{pass}$ and $\textit{score} \geq 2$}
  \State $s \gets \textsc{ExtractSkill}(\text{record})$ 
  \Statex \graycomment{question template, route, abstracted Gremlin; plan + merge if $\rho = \textsc{plan}$}
  \State $\mathcal{S} \gets \mathcal{S} \cup \{s\}$
\EndIf
\State \textbf{return} $\mathcal{S}$
\end{algorithmic}
\end{algorithm}

\section{Post-Hoc Property Verifier}
\label{app:verifier}

The verifier in §\ref{sec:guards} audits captured Gremlin queries
against four similarity-edge correctness properties. This appendix
gives the full definition of each property, the failure mode it
prevents, and the regex pattern used to detect violations. Notation: 
\(\mathcal{E}_{\text{sim}} =
\{\texttt{similarReason},\allowbreak
\texttt{similarSymptoms},\allowbreak
\texttt{similarRootCause}\}\)
is the set of similarity edge labels, and the within-case pairing $\Phi$ maps each
similarity edge to the within-case edge that brackets it:
\[
\begin{aligned}
\Phi(\texttt{similarReason}) &= \texttt{hasReason}, \\
\Phi(\texttt{similarSymptoms}) &= \texttt{exhibits}, \\
\Phi(\texttt{similarRootCause}) &= \texttt{attributedTo}.
\end{aligned}
\]
The verifier walks the captured Gremlin history from each run; a run
record is \emph{verifier-pass} iff all four properties hold on every
captured query. Properties $P_1$–$P_4$ are vacuously \textsc{pass} on
queries that contain no similarity edge step (e.g., Direct route
queries that filter by property only).

\subsection*{P1: Window-on-Destination}

\paragraph{Rationale.} Our evaluation graph is a fixed snapshot of the production support graph with all cases tagged by a \texttt{createdAt} timestamp, and we restrict every evaluation query to a fixed experiment window $[\textsc{WIN\_LO}, \textsc{WIN\_HI}]$ so that ground-truth counts and aggregates are reproducible across runs. The verifier therefore audits where in a similarity walk the window predicate appears: P1 enforces the implementation choice that the window must scope the \emph{peer} cases returned by the similarity walk, not just the anchor case the walk starts from.

\paragraph{Statement.} Every query containing a similarity-edge step
must apply the time-window predicate
$\texttt{P.between}(\textsc{WIN\_LO}, \textsc{WIN\_HI})$ \emph{after} the
post-similarity \texttt{.in($\Phi(e)$)} step that returns to a case
vertex, not on the anchor.

\paragraph{Failure mode prevented.} Applying the window filter on
the anchor case (before the similarity hop) restricts the source
case to the experiment window, but lets the walk return peers from
\emph{any} time period. Conversely, applying it on the destination
case restricts the peer set itself, which is the intended scoping
for similarity questions over the experiment window.

\paragraph{Detection.} Pass iff (a) the query contains
$\texttt{P.between}(\textsc{WIN\_LO}, \textsc{WIN\_HI})$ at least
once, and (b) at least one \texttt{.in($\Phi(e)$)} step is
followed by the window predicate later in the query string.
Queries that have no \texttt{.in($\Phi(e)$)} step (e.g., walks that
end at the entity vertex without returning to a case) are vacuously
\textsc{pass}.

\subsection*{P2: Threshold-on-Edge}

\paragraph{Statement.}
For every similarity-edge step
\(\texttt{bothE}(e)\), \(\texttt{outE}(e)\), or \(\texttt{inE}(e)\),
with \(e \in \mathcal{E}_{\text{sim}}\), the query must apply
\(\texttt{.has('similarityScore',}\allowbreak
\texttt{ P.gte(}\theta\texttt{))}\)
between the edge step and the next vertex step
\(\texttt{.otherV()}\), \(\texttt{.inV()}\), or \(\texttt{.outV()}\).

\paragraph{Failure mode prevented.} Without an explicit threshold,
similarity walks return all peers regardless of similarity score,
which dilutes the result set with low-similarity matches. Applying
the threshold \emph{after} the vertex step instead of on the edge
filters the wrong object — vertices don't carry the similarity
score; the edge does.

\paragraph{Detection.} For each similarity-edge step, find the
substring between it and the next vertex step. Pass iff every such
substring contains
$\texttt{.has(`similarityScore', P.gte(...))}$. Edge steps with no
following vertex step (malformed traversals) are vacuously
\textsc{pass} since they do not reach a peer.

\subsection*{P3: Entity-Bracket}

\paragraph{Statement.} Every similarity-edge step with edge
$e \in \mathcal{E}_{\text{sim}}$ must be preceded by a within-case
$\texttt{.out($\Phi(e)$)}$ step earlier in the query.

\paragraph{Failure mode prevented.} Similarity edges connect content
vertices ($\texttt{reason}$, $\texttt{symptoms}$, $\texttt{rootCause}$),
not cases directly. Walking from a case to a similarity edge without
first traversing the corresponding within-case edge is a structural
error: the traversal has no source content vertex on which to
attach the similarity step.

\paragraph{Detection.} For each occurrence of a similarity edge
with label $e$, scan the query string before the edge step and
require that $\texttt{.out($\Phi(e)$)}$ appears at least once. Note
the pairing is per-label: $\texttt{similarReason}$ requires
$\texttt{.out(`hasReason')}$ specifically, not $\texttt{.out(`exhibits')}$
or $\texttt{.out(`attributedTo')}$.

\subsection*{P4: Anchor-by-Id}

\paragraph{Statement.} Case anchors must be addressed by vertex id,
either via $\texttt{g.V(`uuid')}$ or via $\texttt{.hasId(`uuid')}$.
The label-and-property form $\texttt{g.V().has(`case', `id', `uuid')}$
addresses an \texttt{id} property that does not exist on case vertices
and is forbidden.

\paragraph{Failure mode prevented.} Cases in the live graph do not
carry an \texttt{id} property; their identity is the vertex id
itself. Filtering on $\texttt{.has(`case', `id', `uuid')}$ matches
zero vertices and silently returns an empty result, which the LLM
may misinterpret as a real (negative) finding.

\paragraph{Detection.} Reject any query containing the literal
substring pattern
$\texttt{V().has(`case', `id', ...)}$. Other anchoring forms
($\texttt{g.V(`uuid')}$, $\texttt{.hasId}$) are not flagged.

\subsection*{Verifier Output}

For each captured Gremlin query, the verifier emits a per-property
record:\footnote{The field name \texttt{P4\_anchor\_by\_tilde\_id} refers to Gremlin's \texttt{\textasciitilde id} notation for vertex IDs (the meta-id, distinct from a property named \texttt{id}); the leading tilde is dropped from the JSON key for compatibility.}
\begin{lstlisting}[language=json]
{
  "P1_window_on_destination": {"pass": bool, "reason": str | null},
  "P2_threshold_on_edge":     {"pass": bool, "reason": str | null},
  "P3_entity_bracket":        {"pass": bool, "reason": str | null},
  "P4_anchor_by_tilde_id":    {"pass": bool, "reason": str | null}
}
\end{lstlisting}
A failing property surfaces a structured \texttt{reason} string
identifying the offending substring or position. The run-level
\emph{verifier-pass} flag is the conjunction of all per-query
per-property results across the captured Gremlin history.

\section{Detailed Experimental Setup}
\label{app:setup}

\paragraph{Graph Database.} SEGRA queries an enterprise IT support
property graph, backed by a managed graph database and accessed through
Gremlin, at production scale. Each case vertex is connected to
free-text content vertices ($v_{\text{c}_1}, v_{\text{c}_2}, v_{\text{c}_3}$)
covering incident description, observed behavior, and diagnosed cause),
resolution vertices, a multi-level hierarchical taxonomy of case
categories, and context vertices capturing the parties involved and
device/compliance state. Derived similarity-edge labels
$\mathcal{E}_{\text{sim}} = \{e_1, e_2, e_3\}$ connect content vertices
that share semantic content, each carrying a continuous similarity
score used as a threshold for filtering peers (P2 in
Appendix~\ref{app:verifier}). Every case carries a creation timestamp,
used by the experiment-window predicate for reproducibility (P1).
SEGRA only reads from the graph and never modifies it.

\paragraph{Dataset.} The benchmark has 166 questions split between a 10-question bootstrap set (used to seed $\mathcal{S}$ before evaluation) and a 156-question evaluation set. The evaluation set is partitioned by reasoning shape:
\begin{itemize}[leftmargin=*,itemsep=2pt,topsep=2pt]
  \item \textbf{36 single-shot} questions answered by $H_{\textsc{dir}}$ -- filter-and-count, top-$N$ ranking, and aggregate questions over case properties and the taxonomy.
  \item \textbf{20 similarity-anchored} questions answered by $H_{\textsc{sim}}$ -- cross-case similarity questions over derived \texttt{similar*} edges, either anchored to a referenced case or seeded by free-text retrieval.
  \item \textbf{100 multi-shot} questions answered by $H_{\textsc{plan}}$ -- compositional questions that no single Gremlin query can answer.
\end{itemize}
 
The 100 multi-shot questions span 20 categories drawn from a failure-mode analysis of pilot runs. Categories include parallel comparisons (\emph{``compare resolution rate of VPN vs. Outlook''}), top-$N$ analytics with sub-aggregation, set differences across temporal windows, recurrence patterns over taxonomic categories, and similarity-driven aggregations. The category-level distribution drives the curriculum-bootstrap procedure described in §\ref{sec:skills}. We also include examples from our dataset in \cref{tab:examples} with the natural-language question, the Gremlin issued, and the final answer. The Direct example issues a single aggregation query; the Similarity example anchors on a case and walks the bidirectional similarity edge with score and time-window filters; the Multi-shot example is decomposed by the planner into two parallel branches whose results are combined by the merge step.

\begin{table}[h!]
\centering
\small
\begin{tabular}{p{0.95\linewidth}}
\toprule
\textbf{Example for $H_{\textsc{dir}}$} \\
\midrule
\textbf{Question:} How many cases were created in total? \\
\textbf{Gremlin:} \texttt{g.V().hasLabel('case').count()} \\
\textbf{Answer:} 123{,}456 cases were created in the experiment window. \\
\midrule
\textbf{Example for $H_{\textsc{sim}}$} \\
\midrule
\textbf{Question:} How many in-window cases share similar symptoms with case \texttt{4f4ded41-...} above the production threshold? \\
\textbf{Gremlin:} \texttt{g.V('4f4ded41-...').hasLabel('case').\allowbreak{}out('exhibits').bothE('similarSymptoms').\allowbreak{}has('similarityScore', P.gte(0.675)).otherV().\allowbreak{}in('exhibits').hasLabel('case').\allowbreak{}hasId(P.neq('4f4ded41-...')).dedup().count()} \\
\textbf{Answer:} 4 in-window cases share similar symptoms at or above the production threshold ($\geq 0.675$). \\
\midrule
\textbf{Example for $H_{\textsc{plan}}$} \\
\midrule
\textbf{Question:} Which has more cases -- Mac or Windows? \\
\textbf{Plan:} \texttt{parallel\_compare}; two independent branches, merged with \texttt{compare\_ratios}. \\
\textbf{Gremlin 1:} \texttt{g.V().hasLabel('case').where(out(\allowbreak{}'hasTelemetry').has('devicePlatform', 'MAC')).count()} $\to$ 12{,}345. \\
\textbf{Gremlin 2:} \texttt{g.V().hasLabel('case').where(out(\allowbreak{}'hasTelemetry').has('devicePlatform', 'WINDOWS')).count()} $\to$ 123{,}456. \\
\textbf{Answer:} Windows has more cases than Mac. \\
\bottomrule
\end{tabular}
\caption{Example questions and Gremlin queries for different handlers.}
\label{tab:examples}
\end{table}

\paragraph{Judge Rubric.}\label{app:judge}
The LLM judge (§\ref{sec:metrics}) scores each run on five orthogonal dimensions, each on a 0--4 anchor scale, and reports a per-dimension rationale plus an aggregate. The anchor scale is shared across dimensions: \textbf{4} = great, fully meets the dimension with no caveat worth flagging; \textbf{3} = good, solidly meets, at most a cosmetic nit; \textbf{2} = OK, mostly meets with a defensible minor gap; \textbf{1} = poor, significant gap that affects the answer's value; \textbf{0} = failed, does not meet the dimension at all. The aggregate \texttt{overall\_score} is the \emph{floor} of the mean of the five scores; rounding down rather than to nearest errs on the conservative side.

\begin{enumerate}[leftmargin=*,itemsep=2pt,topsep=2pt]
  \item \textbf{Intent fidelity.} Does the executed Gremlin capture what the user actually asked? Looks at vertex labels, filter predicates (issue category/subcategory, time window, status), traversal direction, and whether the right entities are anchored. \textbf{4}: traversal precisely matches the question, all filters justified. \textbf{3}: matches with one inferred-but-reasonable filter. \textbf{2}: right shape but a non-decisive filter is missing or slightly off. \textbf{1}: wrong vertex type, wrong direction, or a filter inversion. \textbf{0}: traversal answers a fundamentally different question.
  \item \textbf{Execution validity.} Did the query run cleanly and return data of the expected shape? \textbf{4}: ran first try, non-empty data of the expected shape. \textbf{3}: ran cleanly with a small retry/relaxation noted in the trace. \textbf{2}: ran but result is partial / had to be retried successfully. \textbf{1}: errored on first try and recovered with a degraded result, or returned data of the wrong shape that still answers the question. \textbf{0}: errored without recovery, returned empty when data was clearly expected, or the shape is unusable. An ``honest empty'' answer that explains why the data is missing earns at most 2.
  \item \textbf{Grounding.} Is the natural-language answer faithful to what the query actually returned? This is the hallucination check, independent of whether the query was correct. \textbf{4}: every numeric, name, and claim is supported by the result. \textbf{3}: supported with a single rounding or wording inconsistency. \textbf{2}: one minor unsupported detail, headline still correct. \textbf{1}: a key claim is unsupported or contradicts the result. \textbf{0}: fabricates entities, counts, or relationships not in the data.
  \item \textbf{Completeness.} Did the answer address every part of the question, including sub-asks (``compare X and Y'', ``list and explain'', ``rank by Z'')? \textbf{4}: every sub-ask answered with explicit comparisons or explanations. \textbf{3}: every sub-ask answered, one is briefer than ideal. \textbf{2}: main ask answered, secondary sub-ask glossed. \textbf{1}: only one of multiple sub-asks addressed. \textbf{0}: answer is on a different topic or refuses without justification. For single-ask questions, the default is 4 unless something obvious is omitted.
  \item \textbf{Usefulness.} Would an analyst actually use this answer to make a decision? Vague hedging is penalised, while honest fallbacks (e.g., ``vector search unavailable, here is the text-search fallback'') are rewarded because the analyst can act on them. \textbf{4}: directly actionable, no follow-up needed. \textbf{3}: actionable with a one-line caveat the answer surfaces. \textbf{2}: useful with a meaningful caveat. \textbf{1}: ambiguous; analyst would need to re-run the query. \textbf{0}: misleading, evasive, or surfaces no actionable signal.
\end{enumerate}

A run is counted as pass when \texttt{overall\_score} $\geq 2$. 
The judge is instructed to reply with a single JSON object containing the five per-dimension records, the aggregate, and the booleans, with no preamble or code fences.

\section{Per-Dimension Breakdown}
\label{app:dim-breakdown}

Table~\ref{tab:dim-breakdown} reports the mean per-dimension GPT-5.4 judge scores that compose the aggregate \texttt{overall\_score} reported in §\ref{sec:results}, for each (system, backbone) cell over the 156-question evaluation set. The five dimensions follow the rubric defined in Appendix~\ref{app:judge}: \emph{Intent} = intent fidelity, \emph{Exec} = execution validity, \emph{Ground} = grounding, \emph{Compl} = completeness, \emph{Use} = usefulness. The aggregate \texttt{overall\_score} reported in Table~\ref{tab:main-results} is the floor of the per-question mean of these five dimensions, so the per-dimension means here will be slightly higher than the aggregate (mean before floor vs.\ floor before mean across questions).

The pattern across systems is consistent. CoT and Few-shot CoT score reasonably on \emph{Intent} (the model usually understands what is being asked) but collapse on every other dimension because, with no tools, queries either fail outright or return a degenerate shape. Few-shot CoT lifts \emph{Intent} above CoT by $0.4$--$0.6$ points on every backbone -- the worked examples teach traversal patterns -- but \emph{Execution} stays under $1.0$, confirming that single-call prompting cannot recover from the schema-shaped errors that tool feedback would catch. The SEGRA architecture pulls \emph{Execution}, \emph{Grounding}, and \emph{Completeness} into the $2$--$3$ range, and adding the skill library lifts \emph{Usefulness} (the most synthesis-heavy dimension) by a further $0.02$--$0.30$ points.

\begin{table*}[t!]
\centering
\small
\begin{adjustbox}{width=\linewidth}
\setlength{\tabcolsep}{4pt}
\begin{tabular}{l ccccc ccccc ccccc}
\toprule
& \multicolumn{5}{c}{\texttt{Small}} & \multicolumn{5}{c}{\texttt{Medium}} & \multicolumn{5}{c}{\texttt{Large}} \\
\cmidrule(lr){2-6} \cmidrule(lr){7-11} \cmidrule(lr){12-16}
System
& Intent & Exec & Ground & Compl & Use
& Intent & Exec & Ground & Compl & Use
& Intent & Exec & Ground & Compl & Use \\
\midrule
CoT                       & 1.29 & 0.08 & 1.19 & 0.06 & 0.25 & 1.73 & 0.21 & 1.36 & 0.08 & 0.22 & 1.59 & 0.08 & 1.54 & 0.08 & 0.24 \\
Few-shot CoT              & \textbf{1.91} & 0.64 & 0.94 & 0.42 & 0.32 & \textbf{2.17} & 0.88 & 1.12 & 0.36 & 0.22 & \textbf{2.08} & 0.99 & 1.05 & 0.36 & 0.22 \\
SEGRA w/o $\mathcal{S}$   & 1.59 & 2.26 & \textbf{1.68} & 2.54 & 1.24 & 1.90 & 3.10 & 2.05 & 3.13 & 1.46 & 1.90 & 3.12 & 2.13 & 3.31 & 1.73 \\
\textbf{SEGRA w/ $\mathcal{S}$} & 1.74 & \textbf{2.71} & 1.62 & \textbf{2.97} & \textbf{1.25} & 2.16 & \textbf{3.11} & \textbf{2.15} & \textbf{3.15} & \textbf{1.75} & 1.99 & 3.13 & \textbf{2.20} & \textbf{3.41} & \textbf{1.76} \\
\bottomrule
\end{tabular}
\end{adjustbox}
\caption{Per-dimension mean GPT-5.4 judge scores (0--4) over the 156-question evaluation set, by system and backbone. Dimensions are defined in Appendix~\ref{app:judge}. Bold marks the highest-scoring system per (backbone, dimension); SEGRA w/ $\mathcal{S}$ attains the highest score on a majority of cells, with the most consistent gains on \emph{completeness} and \emph{usefulness} (which reward end-to-end synthesis rather than just running a query). Note the contrast on \emph{Intent}: Few-shot CoT slightly exceeds SEGRA on this dimension because both reach a ceiling from prompt-level grounding, but its low \emph{Execution} drags the aggregate down.}
\label{tab:dim-breakdown}
\end{table*}

\begin{table*}[t!]
\centering
\small
\setlength{\tabcolsep}{4pt}
\resizebox{\textwidth}{!}{%
\begin{tabular}{lccccccccccccccc}
\toprule
 & \multicolumn{5}{c}{\texttt{Small}} & \multicolumn{5}{c}{\texttt{Medium}} & \multicolumn{5}{c}{\texttt{Large}} \\
\cmidrule(lr){2-6} \cmidrule(lr){7-11} \cmidrule(lr){12-16}
System & I & E & G & C & U & I & E & G & C & U & I & E & G & C & U \\
\midrule
CoT & 1.29 & 0.08 & 1.19 & 0.06 & 0.25 & 1.73 & 0.21 & 1.36 & 0.08 & 0.22 & 1.59 & 0.08 & 1.54 & 0.08 & 0.24 \\
Few-shot CoT & \textbf{1.91} & 0.64 & 0.94 & 0.42 & 0.32 & \textbf{2.17} & 0.88 & 1.12 & 0.36 & 0.22 & \textbf{2.08} & 0.99 & 1.05 & 0.36 & 0.22 \\
CoT + tools (ReAct-style) & 1.62 & 2.10 & 1.51 & 2.08 & 1.13 & 2.04 & 3.06 & 2.05 & 2.84 & 1.59 & 2.04 & 3.11 & \textbf{2.21} & 2.99 & \textbf{1.82} \\
SEGRA without planner & 1.71 & 2.19 & 1.38 & 2.38 & 1.22 & 1.98 & 3.07 & 2.09 & 2.63 & 1.47 & 1.78 & 2.77 & 2.19 & 2.76 & 1.67 \\
SEGRA without skills & 1.59 & 2.26 & \textbf{1.68} & 2.54 & 1.24 & 1.90 & 3.10 & 2.05 & 3.13 & 1.46 & 1.90 & 3.12 & 2.13 & 3.31 & 1.73 \\
SEGRA, planner-only retrieval & 1.58 & 2.57 & 1.52 & 2.74 & \textbf{1.26} & 2.04 & 3.07 & 2.02 & 3.14 & 1.62 & 1.87 & 3.02 & 2.25 & 3.09 & 1.81 \\
\textbf{SEGRA with skills} & 1.74 & \textbf{2.71} & 1.62 & \textbf{2.97} & 1.25 & 2.16 & \textbf{3.11} & \textbf{2.15} & \textbf{3.15} & \textbf{1.75} & 1.99 & \textbf{3.13} & 2.20 & \textbf{3.41} & 1.76 \\
\bottomrule
\end{tabular}%
}
\caption{Seven-system ladder evaluated by three judge models across five dimensions. \textbf{I} = intent fidelity, \textbf{E} = execution validity, \textbf{G} = grounding, \textbf{C} = completeness, and \textbf{U} = usefulness.}
\label{tab:full}
\end{table*}

\subsection{Human-vs-Judge Agreement}
\label{app:human-judge}

The main paper's per-dimension breakdown reports GPT-5.4 judge scores. We additionally collected human annotations on the SEGRA-w/-skills + \texttt{Medium} cell to gauge how the automatic judge compares to expert evaluation. Two annotators independently scored all 156 questions on the same 0--4 anchor scale as the LLM judge, but assigned a single \emph{overall} score per question rather than per-dimension scores -- so this section is restricted to overall-score agreement and does not extend the per-dimension table above. Annotators saw the question, generated Gremlin, executed result, and final answer, but not the judge score.

\paragraph{Inter-annotator agreement.} Before comparing humans against the judge, we measure how the two annotators agree with each other on the same 156 questions. Table~\ref{tab:iaa} reports five complementary metrics, since the 0--4 scale is ordinal: exact agreement (both annotators picked the same integer), within-1-point agreement (they differed by at most one bucket), Cohen's quadratic-weighted $\kappa$ (the standard ordinal-agreement statistic, where $\geq 0.6$ is conventionally read as ``substantial''), Spearman's $\rho$ (rank correlation, robust to one annotator being uniformly harsher than the other), and mean absolute difference (concrete points-apart). Overall, the annotators agree exactly on $47.4\%$ of questions and within one point on $81.4\%$, with a quadratic-weighted $\kappa$ of $0.52$ and a Spearman $\rho$ of $0.47$ -- moderate agreement under standard interpretation, with the bulk of disagreements being one-bucket calls rather than wholesale reversals.

The handler-level breakdown reveals two distinct patterns. On $H_{\textsc{dir}}$, both annotators rate nearly every output as great ($A$-mean $3.75$, $B$-mean $3.89$, within-1 $94.4\%$); the $\kappa$ of $0.13$ is misleadingly low because the score distribution has almost no variance for $\kappa$ to credit. On $H_{\textsc{plan}}$, agreement is moderate-to-substantial ($\kappa = 0.56$, $\rho = 0.50$, within-1 $82\%$) -- the largest sub-sample ($n{=}100$) and the most representative of overall agreement. On $H_{\textsc{sim}}$, agreement is genuinely poor ($\kappa = 0.06$, exact $15\%$, MAD $1.45$): Annotator $A$ is systematically harsh on similarity outputs (mean $2.45$) while Annotator $B$ rates almost all 20 similarity outputs as great (mean $3.90$). 

\begin{table}[t]
\centering
\small
\begin{adjustbox}{width=\linewidth}
\begin{tabular}{l rrrrrr}
\toprule
Handler & $n$ & Exact & $\leq 1$ pt & $\kappa_q$ & $\rho$ & MAD \\
\midrule
$H_{\textsc{dir}}$  &  36 & 80.6\% & 94.4\% & 0.13 & 0.25 & 0.25 \\
$H_{\textsc{sim}}$  &  20 & 15.0\% & 55.0\% & 0.06 & 0.35 & 1.45 \\
$H_{\textsc{plan}}$ & 100 & 42.0\% & 82.0\% & 0.56 & 0.50 & 0.81 \\
\midrule
All                 & 156 & 47.4\% & 81.4\% & 0.52 & 0.47 & 0.76 \\
\bottomrule
\end{tabular}
\end{adjustbox}
\caption{Inter-annotator agreement on SEGRA w/ $\mathcal{S}$, \texttt{Medium} backbone. ``Exact'' is the percentage of questions on which both annotators picked the same integer score; ``$\leq 1$ pt'' allows a one-bucket disagreement. $\kappa_q$ is Cohen's quadratic-weighted kappa, $\rho$ is Spearman rank correlation, and MAD is the mean absolute difference between the two annotators' integer scores. The $H_{\textsc{dir}}$ $\kappa_q$ is depressed by a ceiling effect (both annotators rate nearly all outputs $\geq 3$).}
\label{tab:iaa}
\end{table}

\section{Additional Results}

\subsection{Question Streaming Order}

To measure the sensitivity of question order during evaluation, we ran the 100 planner evaluation questions under five random orderings, each initialized with the same 10 bootstrapped skill library. And the results are illustrated in \cref{tab:randomorder}.

\begin{table}[t]
\centering
\begin{adjustbox}{width=\linewidth}
\begin{tabular}{ccccc}
\toprule
Seed & Mean score (0--3) & $\geq 2$ rate & $\geq 3$ rate & Skills at end \\
\midrule
0 & 2.674 & 96.7\% & 70.7\% & 108 \\
1 & 2.552 & 91.7\% & 63.5\% & 110 \\
2 & 2.544 & 95.6\% & 58.9\% & 109 \\
3 & 2.534 & 89.8\% & 63.6\% & 110 \\
4 & 2.511 & 92.0\% & 59.1\% & 110 \\
\midrule
\textbf{Cross-seed} & \textbf{2.563 $\pm$ 0.064} & \textbf{93.2\% $\pm$ 2.9\%} & \textbf{63.2\% $\pm$ 4.8\%} & \textbf{109.4 $\pm$ 0.9} \\
\bottomrule
\end{tabular}
\end{adjustbox}
\caption{Sensitivity of evaluation to question ordering across five random seeds.}
\label{tab:randomorder}
\end{table}

Across the five orderings, the mean score is 2.563 ± 0.064, the fraction of questions scoring at least 2 is 93.2\% ± 2.9\%, and the fraction scoring 3 is 63.2\% ± 4.8\%. The variation in mean score is approximately 2\% of the full 0–3 scale, indicating limited sensitivity to question order. The learned skill libraries also converge closely. Starting from the same 10 bootstrap skills, each run promotes approximately 100 additional skills and ends with 108–110 skills. The mean pairwise Jaccard similarity is 0.989 when bootstrap skills are included and 0.988 when considering only skills added during evaluation. Overall, these results suggest that both answer quality and the resulting skill library are stable across the tested question orderings. Our current experiments target the intended streaming setting, in which verified experiences can be incorporated as new questions arrive. 

\subsection{More Baseline Comparison}

We expand the evaluation to a seven-system ladder, including prompt-only, few-shot, ReAct-style, no-planner, no-skills, planner-only retrieval, and full SEGRA configurations. The results are illustrated in \cref{tab:full}.

We observed that (1) adding the schema and examples primarily improves intent recognition, but not execution. Few-shot CoT obtains the highest intent score on each backbone but remains below 1.0 on execution because it cannot inspect the graph, execute queries, or recover from errors. (2) adding ReAct-style tool use produces the largest execution improvement over the prompting-only baselines. This confirms that live schema access and iterative execution feedback are critical for text-to-Gremlin generation. (3) explicit planning most consistently improves completeness: full SEGRA exceeds the no-planner variant by 0.59, 0.52, and 0.65 points on the small, medium, and large backbones, respectively. (4) On the largest backbone, ReAct and full SEGRA perform similarly on execution, while SEGRA achieves substantially higher completeness. This suggests that strong models can perform more implicit planning within a general tool loop, whereas explicit decomposition remains valuable for multi-part questions.

\end{document}